\documentclass[10pt,twocolumn,letterpaper]{article}

\usepackage[pagenumbers]{cvpr} 

\usepackage{graphicx}
\usepackage{amsmath}
\usepackage{booktabs}
\usepackage{tabularx}
\usepackage{colortbl}
\usepackage{pifont}
\usepackage{soul}
\usepackage{multirow}
\usepackage{multicol}
\usepackage{stmaryrd}
\usepackage{amssymb}
\usepackage{caption}
\usepackage{graphicx}
\usepackage{booktabs}

\usepackage[accsupp]{axessibility} 

\definecolor{cvprblue}{rgb}{0.21,0.49,0.74}
\usepackage[pagebackref,breaklinks,colorlinks,allcolors=cvprblue]{hyperref}

\def\paperID{1566} 
\def\confName{CVPR}
\def\confYear{2025}

\title{SCAP: Transductive Test-Time Adaptation via Supportive Clique-based Attribute Prompting}

\author {
    Chenyu Zhang\thanks{Equal contribution},
    Kunlun Xu\textsuperscript{\rm *},
    Zichen Liu\textsuperscript{\rm },
    Yuxin Peng\textsuperscript{\rm },
    Jiahuan Zhou\textsuperscript{\rm }\thanks{Corresponding author}    
    \\
    {\small \textsuperscript{\rm } Wangxuan Institute of Computer Technology, Peking University, Beijing 100871, China}\\
    {\tt\small \{2200010814, xkl, lzc20180720\}@stu.pku.edu.cn, \{pengyuxin, jiahuanzhou\}@pku.edu.cn}    
}

\begin{document}
\maketitle
\begin{abstract}
Vision-language models (VLMs) encounter considerable challenges when adapting to domain shifts stemming from changes in data distribution. Test-time adaptation (TTA) has emerged as a promising approach to enhance VLM performance under such conditions. In practice, test data often arrives in batches, leading to increasing interest in the \textit{transductive TTA} setting. However, existing TTA methods primarily focus on individual test samples, overlooking crucial cross-sample correlations within a batch. While recent ViT-based TTA methods have introduced batch-level adaptation, they remain suboptimal for VLMs due to inadequate integration of the text modality. To address these limitations, we propose a novel transductive TTA framework, Supportive Clique-based Attribute Prompting (SCAP), which effectively combines visual and textual information to enhance adaptation by generating fine-grained attribute prompts across test batches. SCAP first forms supportive cliques of test samples in an unsupervised manner based on visual similarity and learns an attribute prompt for each clique, capturing shared attributes critical for adaptation. For each test sample, SCAP aggregates attribute prompts from its associated cliques, providing enriched contextual information. To ensure adaptability over time, we incorporate a retention module that dynamically updates attribute prompts and their associated attributes as new data arrives.  Comprehensive experiments across multiple benchmarks demonstrate that SCAP outperforms existing state-of-the-art methods, significantly advancing VLM generalization under domain shifts. Our code is available at \href{https://github.com/zhoujiahuan1991/CVPR2025-SCAP}{https://github.com/zhoujiahuan1991/CVPR2025-SCAP}.
\end{abstract}    
\section{Introduction}
\label{sec:intro}

\begin{figure}[tb]
    \centering
    \begin{subfigure}[b]{0.48\textwidth}
        \centering
        \includegraphics[width=1\linewidth]{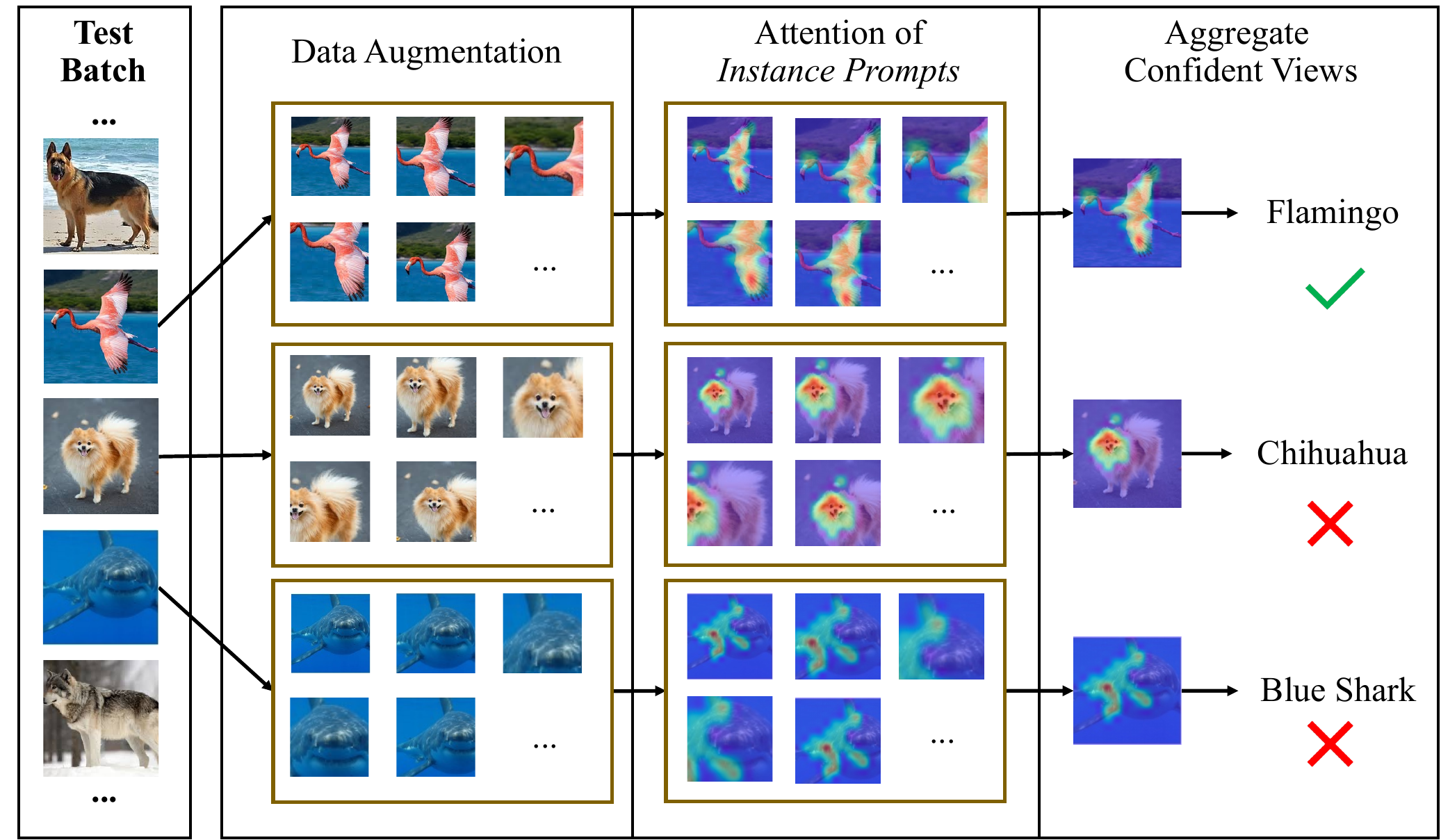}
        \caption{Current Prompt Learning based Test-time Adaptation Methods}
        \label{fig_motive1}
    \end{subfigure}

    \begin{subfigure}[b]{0.48\textwidth}
        \centering
        \includegraphics[width=1\linewidth]{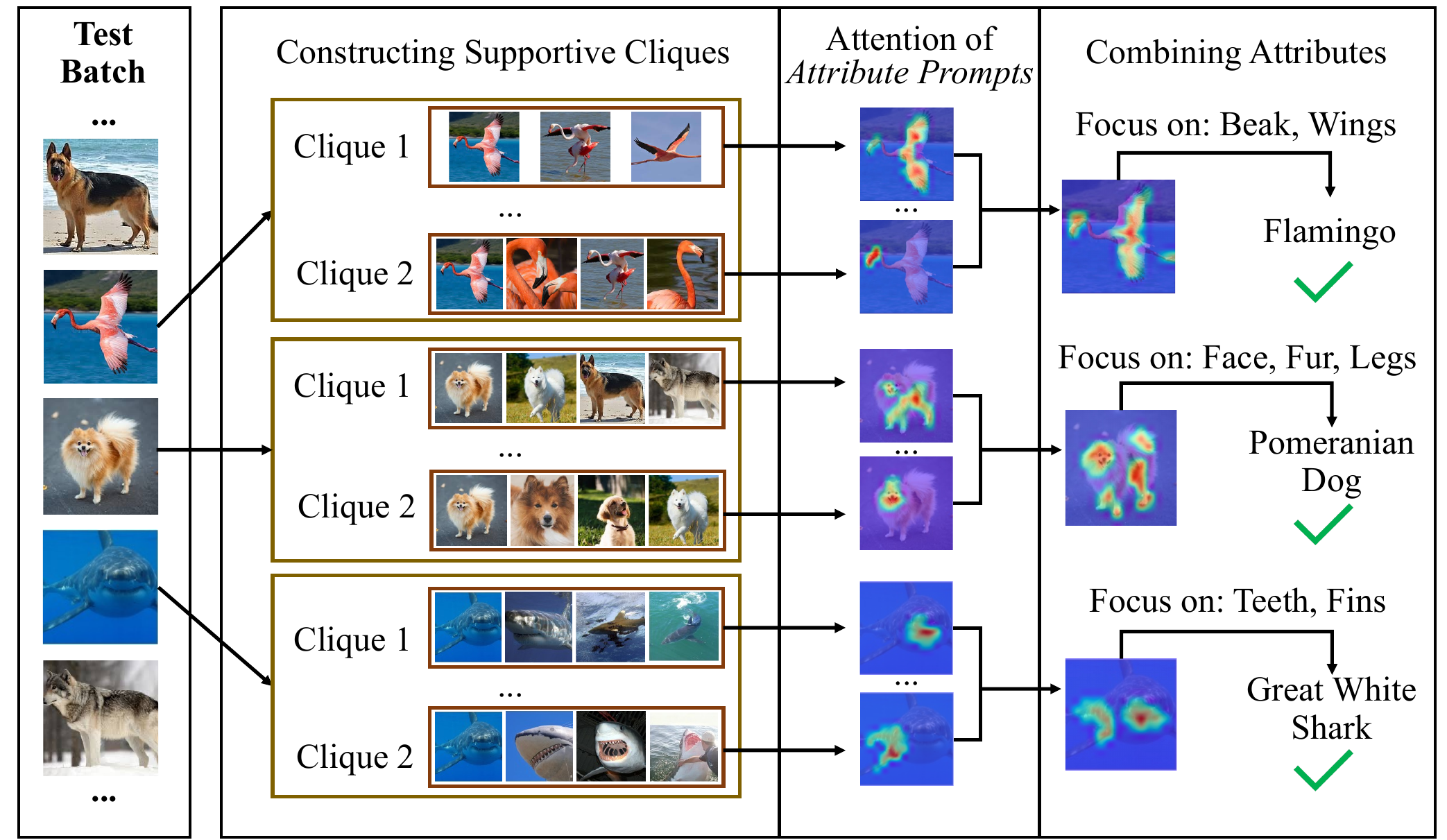}
        \caption{Ours}
        \label{fig_motive2}
    \end{subfigure}
   
    \caption{Comparison between our proposed SCAP to existing prompt learning-based TTA methods. Specifically, current TTA methods learn from instances, while our method utilizes cross-sample visual similarity information from batches to construct \textit{supportive cliques} and extract \textit{attributes} from them. We learn attributes from both modalities based on the cliques. For each image, SCAP jointly utilizes attribute prompts from its associated cliques, leading to more effective and accurate prompting.}
    \label{fig_motive}   
\end{figure}


Vision-language models (VLMs) such as CLIP~\cite{radford2021} have exhibited remarkable generalization capabilities across a range of downstream tasks in zero-shot settings. These advancements highlight the significant potential of CLIP in a wide range of domains.  However, their performance often degrades~\cite{karmanov2024,Liu_Sun_Peng_Zhou_2024,farina2024} when downstream data exhibits substantial domain shifts~\cite{ye2022unsupervised,zhang2024fscil,zhou2020online,xu2024distribution} from the pre-training distribution~\cite{shu2022, zhou2017efficient, fu2022domain,xu2024lstkc}. This limitation has sparked increasing interest in the area of \textit{Test-Time Adaptation (TTA)}~\cite{boudiaf2022parameter,  Zanella_2024_CVPR, chen2022contrastive, wang2020tent}, which aims to adapt pre-trained VLMs to mitigate domain shifts during test time.

Recent works in TTA have primarily focused on adapting CLIP by independently learning from each sample during test time~\cite{shu2022,Zanella_2024_CVPR,Liu_Sun_Peng_Zhou_2024,karmanov2024}. However, in real-world scenarios, test-time data typically arrives in batches, creating a \textit{transductive TTA setting}. Existing TTA methods can tackle such a scenario by treating each test sample individually~\cite{Liu_Sun_Peng_Zhou_2024}. Although this solution is available, it suffers from significant limitations due to disregarding the intrinsic cross-sample visual similarity within each batch. Exploiting these inter-sample similarities can greatly enhance both the effectiveness and robustness of the adaptation process. As illustrated in Figure~\ref{fig_motive}, test batches often contain \textit{Cliques}, groups of visually and semantically similar images that share common characteristics, referred to as \textit{Attributes} in this paper. Leveraging these cliques enables a more precise and context-aware adaptation by guiding the model to focus on discriminative attributes.

Recent works~\cite{sotta2023,wang2021tentfullytesttimeadaptation,press2024rdumb, marsden2023universaltesttimeadaptationweight, hu2021mixnorm} have explored transductive TTA under the Vision Transformer (ViT) backbone, often relying on batch statistics to adjust normalization parameters or select class-balanced samples for adaptation. However, these methods cannot be directly applied to CLIP since they completely neglect the adaptation of the text modality, which is crucial in VLMs.
Rare CLIP-based TTA methods are designed for the transductive setting. In the latest method~\cite{hakim2024clipartt}, the top-k predicted classes of each image in a batch are aggregated into a single new text prompt, which is used as pseudo labels to reclassify inputs in a transductive manner. However, the aforementioned transductive TTA methods mainly perform coarse-grained batch-level utilization of the visual similarity information. Thus, the fine-grained discriminative attribute information of cross-sample visual similarity is underutilized.

In this paper, we propose \textit{Supportive Clique-based Attribute Prompting (SCAP)}, a novel CLIP-based transductive TTA approach designed to effectively leverage the cross-sample visual similarity information. Given a test batch, SCAP automatically mines distinct \textit{Supportive Cliques} containing test images with shared attributes by leveraging visual and semantic similarity metrics. These cliques play a crucial role in learning \textit{Attribute Prompts} that capture discriminative, attribute-specific information for each clique. For any given test sample, its aggregated attribute representation is constructed by combining the learned attribute prompts associated with the cliques to which it belongs. By utilizing these aggregated attribute prompts, SCAP enables CLIP to focus on fine-grained, instance-specific details, enhancing prediction accuracy and robustness. Additionally, we introduce a retention mechanism that dynamically updates and preserves attribute prompts, ensuring long-term adaptability across tasks. Extensive experiments on two benchmarks demonstrate that SCAP significantly outperforms various current state-of-the-art methods in the transductive TTA setting, showing the effectiveness of SCAP in mining the intrinsic knowledge in batches.

    In summary, the contributions of this work are fourfold: (1) 
    We propose Supportive Clique-based Prompting (SCAP), a novel approach that leverages the rich information of cross-sample relationships, improving the effectiveness of test-time adaptation under \textit{transductive setting}. (2) To achieve robust and accurate fine-grained attribute prompt learning during test time, we propose to learn attribute prompts in a transductive manner based on cliques. (3) To retain the historically learned attribute prompts, we further design a retention mechanism to preserve the learned prompts for future reuse. (4) Extensive experiments on two benchmarks demonstrate that SCAP outperforms several current state-of-the-art methods in the transductive TTA setting, highlighting its effectiveness in extracting intrinsic knowledge from batches.

\section{Related Work}
\label{sec:relatedwork}

\subsection{Prompt Learning for CLIP}
VLMs like CLIP have demonstrated abundant knowledge learned from large-scale image-text data~\cite{radford2021, scalingvisualvisionlanguagerepresentation, radford2021learningtransferablevisualmodels, yang2022visionlanguagepretrainingtriplecontrastive, desai2021virtexlearningvisualrepresentations, hu2022scalingvisionlanguagepretrainingimage}. Adapting CLIP to various downstream tasks is a valuable and challenging problem~\cite{cui2024continual}. Recently, prompt learning~\cite{zhang2022tip, yu2023taskresidualtuningvisionlanguage, song2023ecottamemoryefficientcontinualtesttime, li2023graphadaptertuningvisionlanguagemodels, liu2024insvp} has emerged as a promising approach for enhancing model adaptability. Specifically, CoOp \cite{zhou2022learning}, CoCoOp \cite{zhou2022conditional}, MaPLe \cite{maple}, and Tip-Adapter \cite{zhang2022tip} have achieved state-of-the-art performance. However, they rely on training data from downstream tasks. However, these approaches rely on training data annotations and often struggle when faced with substantial domain shifts at test time~\cite{wang2021tentfullytesttimeadaptation,xu2024mitigate}. 


\begin{figure*}[!h]   
  \centering
  \includegraphics[width=\linewidth]{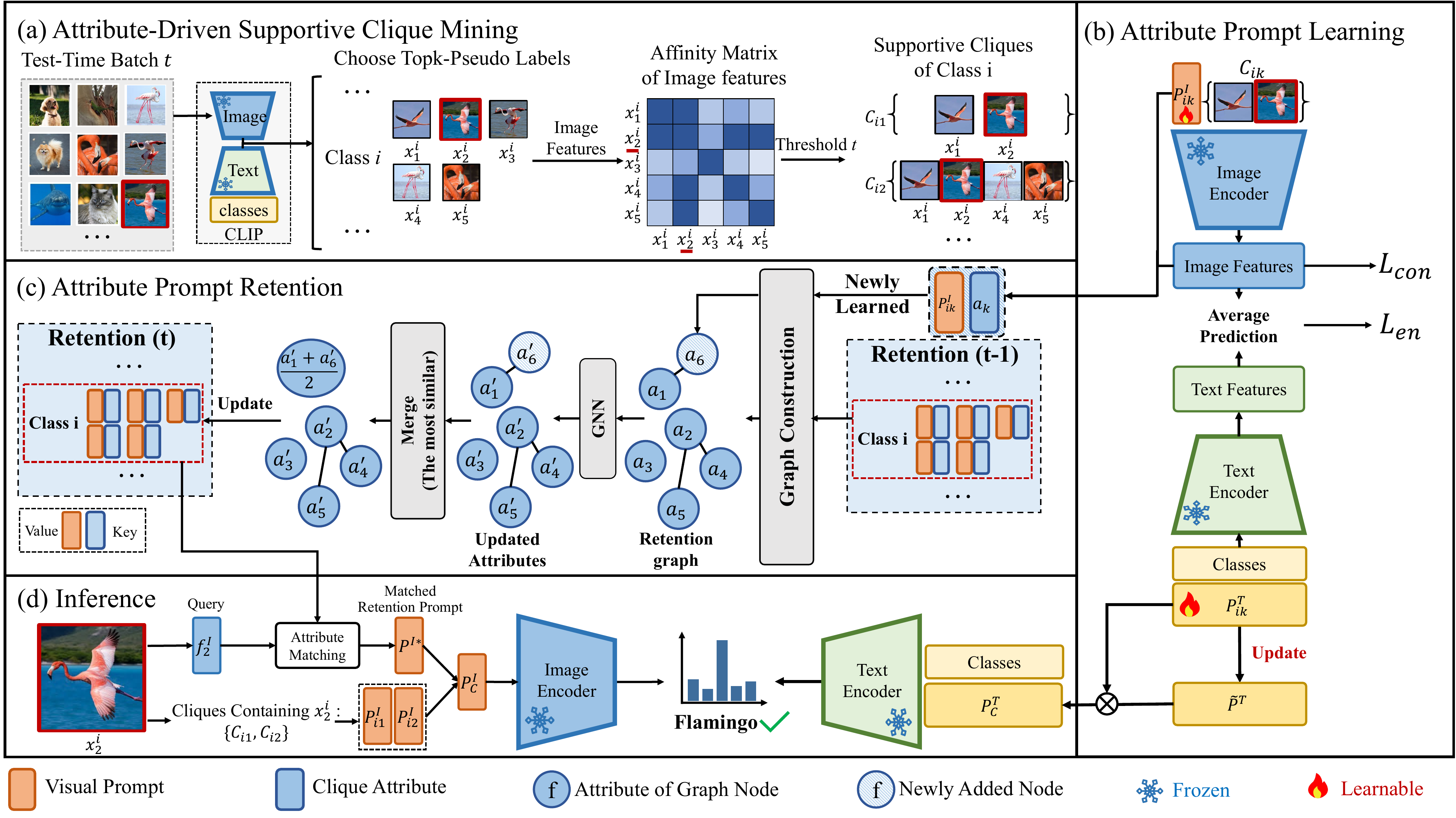}
  \caption{
  Overview of our proposed SCAP. SCAP firstly mines the \textit{supportive cliques} for all images in parallel. Based on the cliques, it then learns the corresponding \textit{attribute prompts} from both modalities. The learned \textit{visual attribute prompts} and the \textit{text attribute prompts} are separately retained to accumulate the knowledge of test domains. For each instance, we conduct inference by jointly utilizing all the associated attribute prompts and the retained knowledge to generate the final prediction.
  }
  \label{fig_arch}
\end{figure*}

\subsection{Test-Time Adaptation}
Test-time adaptation (TTA) focuses on dynamically adjusting pre-trained models, such as CLIP or ViTs~\cite{Liu_Sun_Peng_Zhou_2024,li2024exemplar}, by leveraging sequentially obtained unlabeled test data~\cite{Adaprompt, decoratenewcomersvisualdomain, wang2021tentfullytesttimeadaptation, Zanella_2024_CVPR, feng2023diversedataaugmentationdiffusions}. 
Existing TTA approaches can be broadly categorized into training-free, augmentation-based, and prompt-based methods. 
Training-free methods aim to adapt CLIP efficiently without backpropagation~\cite{karmanov2024, farina2024}. TDA introduces a positive and negative cache mechanism to assist prediction. Zero~\cite{farina2024} aggregates predictions from multiple augmented versions without modifying model parameters.
The augmentation-based methods \cite{roid, Zanella_2024_CVPR} focus on generating diverse augmented samples to integrate test distribution knowledge into the model.
MTA \cite{Zanella_2024_CVPR} introduces a quality assessment variable for each augmented view and models the distribution using a mixed Gaussian approach. 
Prompt-based methods employ prompt tuning to adapt the model while preserving generalization capabilities~\cite{Adaprompt, shu2022, Liu_Sun_Peng_Zhou_2024, ma2024swapprompt, gan2023decorate, yoon2024c}. For example, TPT \cite{shu2022} optimizes instance-specific text prompts by minimizing prediction entropy, whereas DART~\cite{Liu_Sun_Peng_Zhou_2024} refines prompts across both modalities while leveraging historically learned knowledge via prompt retention.
While these methods assume a sequential data setting where test samples are processed individually~\cite{wang2021tentfullytesttimeadaptation, sotta2023}, real-world scenarios often involve test-time data arriving in batches. Consequently, existing approaches struggle to exploit the inter-sample relationships within a batch, leading to suboptimal adaptation performance.

\subsection{Transductive Test-Time Adaptation}
\textit{Transductive TTA} considers the practical scenario where test data are provided in batches rather than individual samples. Existing Transductive TTA methods are predominantly developed for ViTs pre-trained on ImageNet~\cite{sotta2023, wang2020tent, marsden2023universaltesttimeadaptationweight, schneider2020improving, wang2022continual, press2024rdumb}. These approaches typically rely on batch statistics to update normalization parameters~\cite{wang2020tent} or select confident samples for model tuning~\cite{marsden2023universaltesttimeadaptationweight}. 
Recent studies have demonstrated that CLIP outperforms ViTs pre-trained on ImageNet in TTA tasks. However, existing transductive TTA methods fail to fully exploit CLIP’s capabilities, as they largely disregard the adaptation of the text modality, which is crucial for leveraging its knowledge learned from vision-language alignment.
A few recent efforts have attempted to adapt CLIP for transductive TTA. For example, CLIPArTT~\cite{hakim2024clipartt} optimizes the model by jointly utilizing all images within a batch. However, since cross-instance relationships within a batch are often imbalanced, directly modeling all samples together does not effectively exploit the most informative relationships. In this work, we take a step further by identifying and leveraging the informative cliques, groups of strongly correlated samples within a batch, to facilitate robust adaptation by capturing discriminative attributes.

\section{Method}

\subsection{Problem Setting and Notations}

In transductive TTA, test data arrive sequentially in batches. 
Given $t$-th test batch $B_t = \{x_j\}_{j=1}^b$, where $b$ is the batch size and $x_j$ is a test image. We generate a set of Supportive Cliques for each class $i$, denoted as $\mathcal{C}_i = \{ C_{ik}\}_{k=1}^m$. Then, we extract the corresponding clique attributes, which are denoted as $\mathcal{A}_i = \{a_{ik}\in\mathbb{R}^d\}_{k=1}^m$ where $d$ is the attribute dimension, equivalent to the feature dimension. Additionally, we introduce Attribute Prompts $\mathcal{P}_i^A = \{\mathcal{P}_i^I, \mathcal{P}_i^T\} $, which consist of visual attribute prompts $\mathcal{P}_i^I = \{P_{ik}^I\}_{k=1}^m$ and text attribute prompts $\mathcal{P}_i^T = \{P_{ik}^T\}_{k=1}^m$. 
Our model $\theta$ is built on CLIP, which comprises an image encoder $\textbf{E}_I$ and a text encoder $\textbf{E}_T$.

\subsection{Attribute-Driven Supportive Clique Mining}

Attributes are the fine-grained characteristics shared by a group of images in the batch. Since images containing the same attribute often exhibit high visual similarity to each other, in this section, we propose to obtain Supportive Cliques for the potential attributes by evaluating cross-image similarity.

Specifically, we use CLIP to obtain the top-\( k \) class predictions $topk(x_j)$ for each image $x_j$, where \( k \) is a hyperparameter. For each class \( i \), all images whose top-\( k \) predictions include class \( i \) are selected to form a subset \( S_i \):
\begin{equation}
    S_i = \{x_j \in B_t : i \in topk(x_j)\}
    \label{topk}
\end{equation}

Since images in $S_i$ share similar class-specific information, which ensures semantic relevance. To further capture more fine-grained relevance, we extract the features $F^I_i\in \mathbb{R}^{|S_i|\times d}$ of images within $S_i$ using the CLIP image encoder $\textbf{E}_I$ and compute the similarity matrix \( M_i \in \mathbb{R}^{|S_i|\times |S_i|}\) as follows:
\begin{equation}
    M_i = F^I_i \cdot {(F^I_i)}^\top
\end{equation}
Then, given the $l$-th row of \( M_i \), we collect elements exceeding a threshold $t$, whose corresponding images form a Clique \( C_{il} \).  Formally, the Clique generated from the $l$-th row of \( M_i \) is defined as:

\begin{equation}
    C_{il} = \{x_j^i : x_j^i \in S_i, (M_{i})_{lj} > t\},
\end{equation}
where $(M_{i})_{lj}$ represents the $l$-th row, $j$-th column of $M_i$.
Only non-trivial Cliques (\( |C_{il}| > 1 \))are retained, and duplicate Cliques are removed to ensure uniqueness. Finally, we obtain a set of valid Cliques $\mathcal{C}_i = \{ C_{ik}\}_{k=1}^m$.

\subsection{Attribute Prompt Learning}\label{sec:attri_prompt_learn}
In this section, we aim to learn prompts that capture attribute-specific knowledge to facilitate classification.

Specifically, for each Clique $C_{ik}$, we learn a visual attribute prompt $P_{ik}^A$ from the image modality and a text attribute prompt $P_{ik}^T$ from the text modality. 
The feature $f^I_j\in\mathbb{R}^d$ for the $j$-th image in $C_{ik}$ is extracted as:
\begin{equation}
f^I_j = \textbf{E}_I(C_{ik}^j, P^I_k)
\end{equation}
Then, we define the attribute feature $a_{ik}\in\mathbb{R}^d$ of $C_{ik}$ which is calculated by:
\begin{equation}
    a_{ik} = \frac{1}{|C_{ik}|}\sum_{j=1}^{|C_{ik}|}f_j^I
\end{equation}
Besides, we compute the text feature $f^T_n$ of the $n$-th class as follows:

\begin{equation}
f^T_n = \textbf{E}_T(c_n, P^T_{ik})
\end{equation}

Given total $N$ classes, we introduce an attribute-specific class probability distribution vector, which is obtained by:
\begin{equation}
    \hat{\textbf{P}}_{ik} = [\hat{p}_{ik}^1, \hat{p}_{ik}^2, \dots, \hat{p}_{ik}^N]
    \label{pred}
\end{equation}
where
\begin{equation}
    \hat{p}_{ik}^{n} = \frac{\exp \left( \cos(a_{ik}, f_n^T) / \tau \right)}{\sum_{j=1}^{N} \exp \left( \cos(a_{ik}, f_{j}^T) / \tau \right)}
\end{equation}
Note that $\cos(\cdot, \cdot)$ denotes cosine similarity and $\tau$ is a temperature parameter to scale the similarity scores.

Then, we compute the \textit{Entropy Loss} for the probability distribution of the attribute:
\begin{equation}
    \mathcal{L}_{en}(C_{ik}) = -\sum_{n=1}^N \hat{p}_{ik}^{n} \log \hat{p}_{ik}^{n}
    \label{len}
\end{equation}

To further guide the model to extract attribute-specific information and discard attribute-irrelevant information, we introduce a concentration loss $\mathcal{L}_{con}(C_{ik})$ that forces the model to capture the shared information across the samples within a Clique. Specifically, $\mathcal{L}_{con}(C_{ik})$ is calculated by
 \begin{equation}
    \mathcal{L}_{con}(C_{ik}) = \sum_{j=1}^{|C_{ik}|} {|| f_j^I - a_{ik} ||}^2
    \label{lcon}
\end{equation}

Therefore, both kinds of prompts $P^I_{ik}$ and $P^T_{ik}$ are guided to encode attribute-specific knowledge as the model optimization. Such knowledge is valuable when generating predictions for images showing the same attributes.

\subsection{Attribute Prompt Retention}
In this section, to retain attribute knowledge during the test process and ensure consistent performance improvement, we introduce mechanisms for retaining both text and visual attribute knowledge.

Firstly, to retain the text attribute knowledge, we propose a  \textit{Text Retention Prompt} $\tilde{P}^T$. Specifically, when a new Clique $C_{ik}$ is learned, we update $\tilde{P}^T$ by
\begin{equation}
    \tilde{P}^T = \alpha_{\tau}\tilde{P}^T + (1-\alpha_{\tau})P_{ik}^T,
\end{equation}
where $\alpha_{\tau} = \tau/(1+\tau)$ is a decay rate based on iteration count  $\tau$.

Besides, to retain the visual attribute knowledge, we design a \textit{Retention Cache} $\mathcal{R} = \{R_i\}_{i=1}^N$ where $R_i$ is a class-specific retention cache. Specifically, we store $(a_{ik}, P^I_{ik})$ as key-value pairs in $R_i$. We denote the retained attributes (\ie keys) as $R_i^{A}=\{a_{1}, \dots, a_{L}\}$ and the corresponding attribute prompts (\ie values) as $R_i^{P}=\{P_{1}^I, \dots, P_{L}^I\}$. The maximum size of each $R_i$ is set to $L$ to avoid memory explosion. When the retained key-value pair number is less than $L$, we directly add a new $(a_{ik}, P^I_{ik})$ to $R_i$. Otherwise, we add $(a_{ik}, P^I_{ik})$ to $R_i$ first, then we update all pairs and aggregate the two that are most similar to each other in $R_i$ by using a graph structure as follows.

Firstly, we construct a weighted graph $G_i = (V_i, E_i)$. The vertices are the keys in the retention cache $V_i = R_i^{A} = \{a_{1}, \dots, a_{L+1}\}$. The edges $E_i$ between each two vertices are defined as the Gaussian Distance of the two attributes. Thus, the adjacency matrix $W_i\in\mathbb{R}^{(L+1)\times (L+1)}$ of $G_i$ is defined with:
\begin{equation}
    (W_i)_{kl} = exp(-\frac{{\|a_k - a_l\|}^2}{2\sigma^2}),
\end{equation}
where $(W_i)_{kl}$ is the $k$-th row, $l$-column of $W_i$ and $\sigma$ is a hyperparameter.
We utilize the $K$-max values in each row of $W_i$ to construct a $K$-nearest retention graph. Then we perform graph propagation following \cite{graph} and obtain the \textit{updated} graph vertices $V_i' = \{a'_1, \dots, a_{L+1}'\}$ , where the similar attributes are pushed closer automatically. Then, the most similar two attributes $a_{j_0}', a_{j_1}'$ and the corresponding prompts are fused into one respectively as follows:
\begin{equation}
\left\{
		\begin{aligned}
			&  a_{new} = \frac{1}{2}(a_{j_0}' + a_{j_1}')\\
			& P^I_{new} = \frac{1}{2}(P^I_{j_0} + P^I_{j_1})
		\end{aligned}
		\right.
\end{equation}

Through the above process, we obtain the \textit{updated} retention cache $R_i$, dynamically retaining the visual and text attribute knowledge learned from previous batches into \textit{Retention Cache} and \textit{Text Retention Prompt}, respectively. 


\subsection{Optimization and Inference}
\label{training}

For each batch $B_t$, we optimize the visual attribute prompts  $\mathcal{P}_i^I = \{P_{ik}^I\}_{k=1}^m$ and text attribute prompts $\mathcal{P}_i^T = \{P_{ik}^T\}_{k=1}^m$ for every class $i$ by minimizing the following loss function:
\begin{equation}
    \mathcal{L}= \sum_{k=1}^m \mathcal{L}_{en}(C_{ik}) + \lambda \mathcal{L}_{con}(C_{ik})
\end{equation}
where $\lambda$ is a hyperparameter. 

Then we conduct inference for each $S_i$ of $B_t$. To fully utilize both the \textit{Attribute Prompts} learned from the current batch $B_t$ and the \textit{retained knowledge} to improve prediction. Given the test image $x_j \in S_i$, we first collect all Cliques containing $x_j$, namely $\mathcal{K} = \{k: x_j \in C_{ik}\}$, whose corresponding visual attribute prompts are concatenated together:
\begin{equation}
    P^I = concat([P_{ik}^I: k \in \mathcal{K}])
    \label{concat1}
\end{equation}

Then, we introduce the retention prompt matching strategy utilizing the key-value pair structure of the retention cache. Specifically, the keys are the retained attribute features $R_i^{A}=\{a_{1}, \dots, a_{L}\}$ and the values are the corresponding attribute prompts $R_i^{P}=\{P_{1}^I, \dots, P_{L}^I\}$. Given the image feature $f_j^I$ of $x_j$ which is already extracted in \cref{sec:attri_prompt_learn}, the top-1 attribute $a^*$ is obtained by evaluating the similarity of $f_j^I$ to all keys in $R_i^{A}$, \ie
\begin{equation}
    a^* = \underset{a_l\in R_i^{A}}{\mathrm{argmax}} <{f_j^I}, a_l>
\end{equation}
Given the matched value corresponding to $a^*$, \ie, $P^{I*}$, we obtain the \textit{Composed Image Prompt} $P_C^I$ by concatenating $P^I$ and $P^{I*}$ along the token dimension
\begin{equation}
 P_C^I = concat(P^I, P^{I*})    
 \label{concat2}
\end{equation}

For the text modality, we fuse $\mathcal{P^T} = \{P_{ik}^T\}_{k=1}^m$ and $\tilde{P}^T$ together to get the \textit{Composed Text Prompt} $P_C^T$:
\begin{equation}
    P_C^T = \alpha_r \tilde{P}^T + (1-\alpha_r) \frac{1}{m} \sum_{k=1}^m P_{ik}^T
\end{equation}
where $\alpha_r$ is the weight of retained text prompts.
Then, for the final prediction, the image-inference feature $\hat{f}_j^{I}$ of $x_j^i$ and the text-inference feature $\hat{f}_n^{T}$ of class $c_n$ are obtained by:
\begin{equation}
\left\{
		\begin{aligned}
			&  \hat{f}^I_j = \textbf{E}_I(x_{j}, P^I_C) \\
			& \hat{f}^T_n = \textbf{E}_T(c_n, P^T_C)
		\end{aligned}
		\right.
\end{equation}

The prediction $y_j$ of $x_j$ is obtained by applying Eq.\ref{pred}.

\begin{table*}[ht]
  \centering
  \setlength{\tabcolsep}{2.4mm}{
  \begin{tabular}{l l *{7}{c} } 
    \toprule
    {Method}      & Publication   &  ImageNet-A  &  ImageNet-R  & ImageNet-S
    & ImageNet-V2 
    &{Average}    \\
  \midrule
  CLIP  \cite{radford2021}     &  ICML 2021      &	47.87	&73.98	&46.09	&61.88	&57.46       \\
  \midrule
  CoOp \cite{zhou2022learning}       &  IJCV 2022      & 49.71    &75.21  &47.99  &64.20  &59.28 \\
  CoCoOp \cite{zhou2022conditional}    &  CVPR 2022      & 50.63    &76.18  &48.75  &64.07  &59.91 \\
  Tip-Adapter \cite{zhang2022tip}&  ECCV 2022      & 51.04    &77.76  &48.88  &63.41  &60.27 \\
  \midrule
  TPT \cite{shu2022}   &  NeurIPS 2022    & 54.77	&77.06	&47.94	&63.45	&60.81   \\
  C-TPT \cite{yoon2024c}   &  ICLR 2024      & 52.90    &78.00  &48.50  &63.40  &60.70  \\
  DART \cite{Liu_Sun_Peng_Zhou_2024}       &  AAAI 2024     & 60.56	&79.56	&49.76	&64.03	&63.48 \\
  MTA \cite{Zanella_2024_CVPR}   &  CVPR 2024      & 58.06	&78.33	&49.61	&64.24	&62.56\\
  TDA \cite{karmanov2024}       &  CVPR 2024      & 60.11	&80.24	&50.54	&\textbf{64.67}	&63.89\\
  Zero  \cite{farina2024}     &  NeuIPS 2024    & 61.35	&77.28	&48.29	&64.13	&62.76   \\
  \hline
  \rowcolor[HTML]{ECF5F9}
  \textbf{SCAP}   &  \textbf{This Paper}  & \textbf{64.52}	& \textbf{81.68} & \textbf{51.65} & 64.65 & \textbf{65.63} \\
  \noalign{\hrule height 0.295mm}
  \end{tabular}
}
\caption{\textbf{Results comparison on the out-of-distribution (OOD) Benchmark}. The results of the compared methods follow the original paper or are implemented with the official code~\cite{Liu_Sun_Peng_Zhou_2024}.
}
\label{tab:ood-main}
\end{table*}

\section{Experiments}
\begin{figure*}[!h]   
  \centering
  \includegraphics[width=\linewidth]{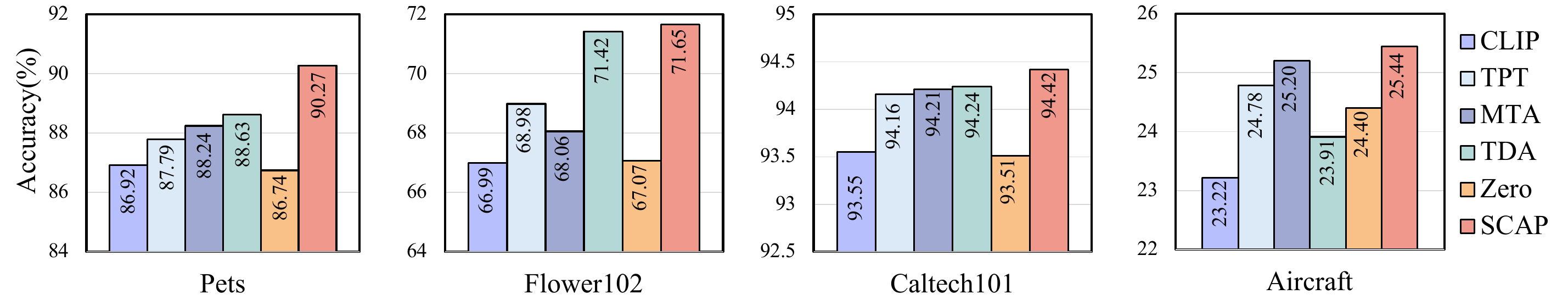}
  \caption{\textbf{Results on the Cross-Domain Benchmark.} Comparison of SCAP with state-of-the-art methods. All methods are evaluated using CLIP-ViT-B/16 as the backbone.
  }
  \label{tab:crossdomain}  
\end{figure*}

 In this section, we first introduce the two benchmarks used for model evaluation, followed by a detailed description of the implementation setup. Next, we present a comparative analysis of existing methods and our proposed approach. Finally, extensive ablation studies are conducted to validate the effectiveness of SCAP. Additional experimental results and in-depth discussions are provided in the Supplementary Materials.

\subsection{Benchmarks}
Our evaluation is conducted on two benchmarks: an out-of-distribution (OOD) benchmark and a cross-domain benchmark. The OOD benchmark consists of four datasets derived from ImageNet \cite{deng2009imagenet}, \ie, ImageNet-A \cite{hendrycks2021natural}, ImageNet-R \cite{hendrycks2021facesrobustnesscriticalanalysis}, ImageNet-Sketch \cite{wang2019learningrobustglobalrepresentations}, and ImageNet-V2 \cite{recht2019imagenetclassifiersgeneralizeimagenet}. The OOD benchmark aims to assess the model’s robustness to substantial distribution shifts in the test data. 
The cross-domain benchmark comprises four datasets of the image classification task, each representing a distinct domain: Aircraft \cite{maji2013finegrainedvisualclassificationaircraft}, Caltech101 \cite{fei2004learning}, Flower102 \cite{nilsback2008automated}, and Pets \cite{parkhi2012cats}. The cross-domain benchmark provides a measure of the model's adaptability across a diverse range of domain-specific tasks.

\paragraph{Implementation details.} 
We adopt CLIP with ViT-B/16 as the backbone and set the batch size to 64 for all experiments. Visual attribute prompts are initialized using a uniform distribution in the range (-1, 1), following previous TTA methods~\cite{Liu_Sun_Peng_Zhou_2024, shu2022}. For text attribute prompts, we employ an ensemble of multiple templates for the OOD benchmark, whereas dataset-specific templates are used for the cross-domain benchmark, following TDA~\cite{karmanov2024}. Our key hyperparameters include the threshold $t$ for supportive clique extraction, the weight $\alpha_r$ for computing $P_C^T$, the concentration loss weight $\lambda$, and graph structure parameters $\sigma$, $L$, and $K$. The hyperparameter ablations of this paper are conducted on ImageNet-A, with the default values of $\alpha_r$, $\sigma$, $L$, and $K$ set to 1, 0.3, 6, and 3, respectively. We set $k=3$ in the initial top-k predictions \cref{topk}. An Adam optimizer with a learning rate of 0.003 is used to update the prompts, and all experiments are conducted on a single NVIDIA 4090 GPU.

\subsection{Comparison with State-of-the-arts}
\textbf{Compared methods.} We compare our proposed SCAP with state-of-the-art TTA methods designed for CLIP adaptation, including TPT~\cite{shu2022}, 
C-TPT~\cite{yoon2024c}, DART~\cite{Liu_Sun_Peng_Zhou_2024}, MTA~\cite{Zanella_2024_CVPR}, TDA~\cite{karmanov2024}, and Zero~\cite{farina2024}. Additionally, we report results from the original CLIP model to highlight the knowledge gained during testing. Following previous works~\cite{karmanov2024}, we also include training-time adaptation methods such as CoOp~\cite{zhou2022learning}, CoCoOp~\cite{zhou2022conditional}, and Tip-Adapter~\cite{zhang2022tip} for comparison to further verify the label efficiency and effectiveness of our approach.

\textbf{Results on the OOD Benchmark.} 
As shown in \cref{tab:ood-main}, our proposed SCAP achieves state-of-the-art performance on the ImageNet-A, ImageNet-R, and ImageNet-S datasets, outperforming existing methods by at least \textbf{3.17\%}, \textbf{1.44\%}, and \textbf{1.11\%}, respectively. Additionally, on the challenging ImageNet-V2 dataset, SCAP performs comparably to TDA, with only a marginal 0.02\% difference. While TDA benefits from a feature cache to enable joint learning across multiple samples, it incurs additional storage overhead. In contrast, SCAP achieves comparable or superior performance without maintaining historical features, making it more efficient in practice.

Overall, SCAP improves average accuracy across the four ImageNet variants by \textbf{1.74\%}, highlighting its effectiveness in handling natural object images across diverse distribution shifts. This performance gain is primarily attributed to the proposed supportive clique mining strategy, which effectively captures diverse object attributes, such as ``head" and ``body", that enhance the object recognition capacity of the model.

\textbf{Results over cross-domain benchmark.}
As shown in~\cref{tab:crossdomain}, we compare our SCAP method with state-of-the-art approaches across four cross-domain datasets. The results demonstrate that our method surpasses the state-of-the-art approaches, with improvements of at least \textbf{1.64\%}, \textbf{0.23\%}, \textbf{0.18\%}, and \textbf{0.24\%} on the Pets, Flower102, Caltech101, and Aircraft datasets, respectively. This superior performance stems from the proposed Attribute Prompt Learning and Attribute Prompt Retention mechanisms, which effectively capture and leverage generalizable knowledge shared across diverse samples.

\begin{table}[tb]
\centering
    \begin{tabularx}{8cm}{>{\centering\arraybackslash}X|>{\centering\arraybackslash}X|>{\centering\arraybackslash}X|c}
    \hline
    \multicolumn{3}{c|}{Components in SCAP} & ImageNet-A \\
    \hline
    Image & Text & Retention & Acc@1  \\
    \hline
    \ding{55} & \ding{55} & \ding{55} & 60.98 \\
    \ding{55} & \ding{51} & \ding{55} & 62.57 \\
    \ding{51} & \ding{55} & \ding{55} & 63.14 \\
    \ding{51} & \ding{55} & \ding{51} & 63.82 \\
    \hline
\rowcolor[HTML]{ECF5F9}
    \ding{51} & \ding{51} & \ding{51} & \textbf{64.52}  \\
    \hline
  \end{tabularx}
  \caption{Ablation study about the different components}
    \begin{flushleft}
        \footnotesize{- \textit{Image} and \textit{Text} represent learning from the Image Modality and Text Modality in the \textit{Attribute Prompt Learning} module, respectively.} \\
        \footnotesize{- \textit{Retention} refers only to the \textit{Retention Cache} for the visual modality while the \textit{text retention prompts} are always used.}
    \end{flushleft}
  \label{tab:components}  
\end{table}

\subsection{Ablation Studies}

\textbf{The Influence of Each Component.} 
To evaluate the effectiveness of our module designs across different modalities (\ie, image, text, and retention), we conduct modality-specific ablation studies. As shown in \cref{tab:components}, leveraging both image and text modalities within the \textit{Attribute Prompt Learning} module consistently improves performance over the baseline model. Additionally, incorporating the \textit{Attribute Prompt Retention} module further enhances the utilization of learned image prompts, leading to substantial performance gains. Furthermore, when all modules and modalities are integrated, the model performance improves further due to joint optimization, which mitigates semantic drift as learnable parameters are updated.

\begin{figure}[!t]
\centering
\includegraphics[width=\linewidth]{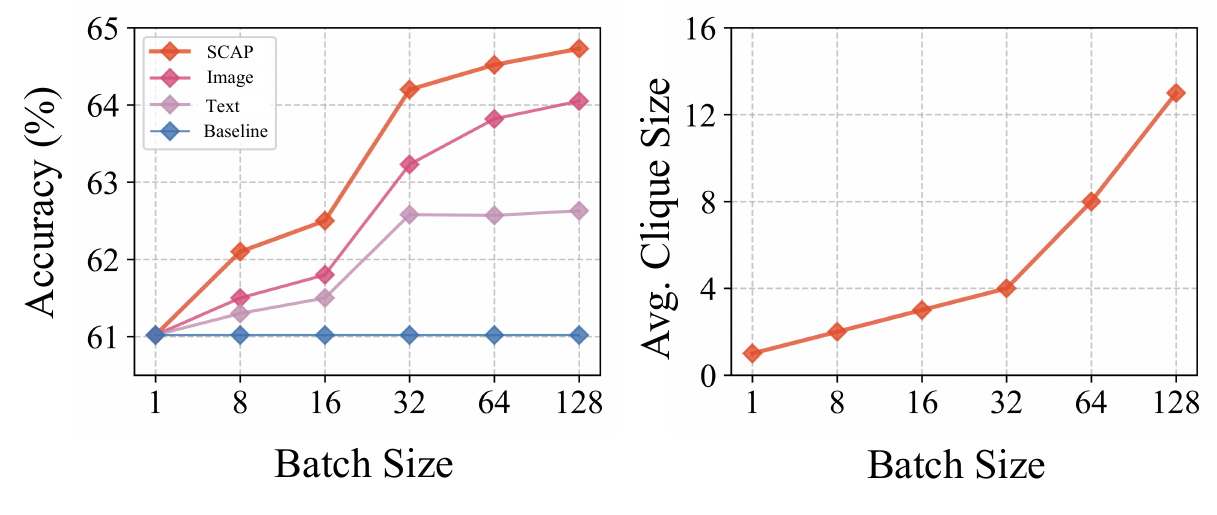}
  \vfill 
  \caption{Left: the influence of batch size on SCAP performance. Right: the average maximum Clique size generated per batch under different batch sizes.}
\label{fig:batchsize}
\end{figure}

\textbf{The Influence of Batch Size in SCAP.} 
 Batch Size $B$ is an important factor in the transductive TTA setting, as larger batches provide richer inter-image relationships. As illustrated in \cref{fig:batchsize} (Left), the model performance consistently improves with increasing batch size, demonstrating SCAP's capability to leverage the additional contextual information. We also observe that the image modality varies significantly as batch size changes, while the text modality remains relatively stable across batch size variations.
On the right side of \cref{fig:batchsize}, the results reveal that larger batches lead to increasing average clique size, indicating that more support information can be captured under such conditions.

\begin{figure}[!t]
\centering
\includegraphics[width=\linewidth]{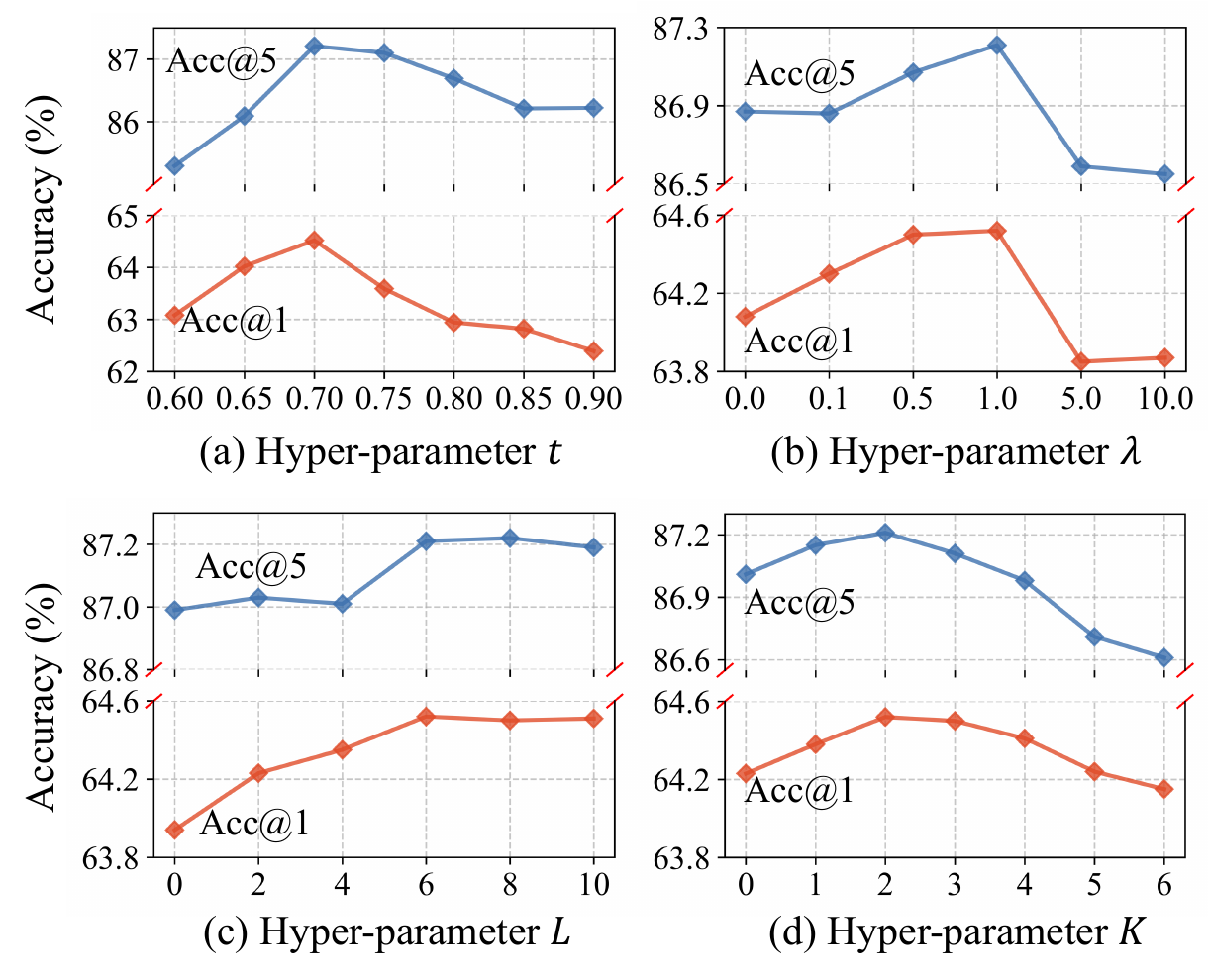}
  \vfill
  \caption{Ablation study about the influence of different hyperparameters on ImageNet-A.}
\label{fig:hyper-parameters}
\end{figure}

\begin{figure*}[!t]
\centering
\includegraphics[width=\textwidth]{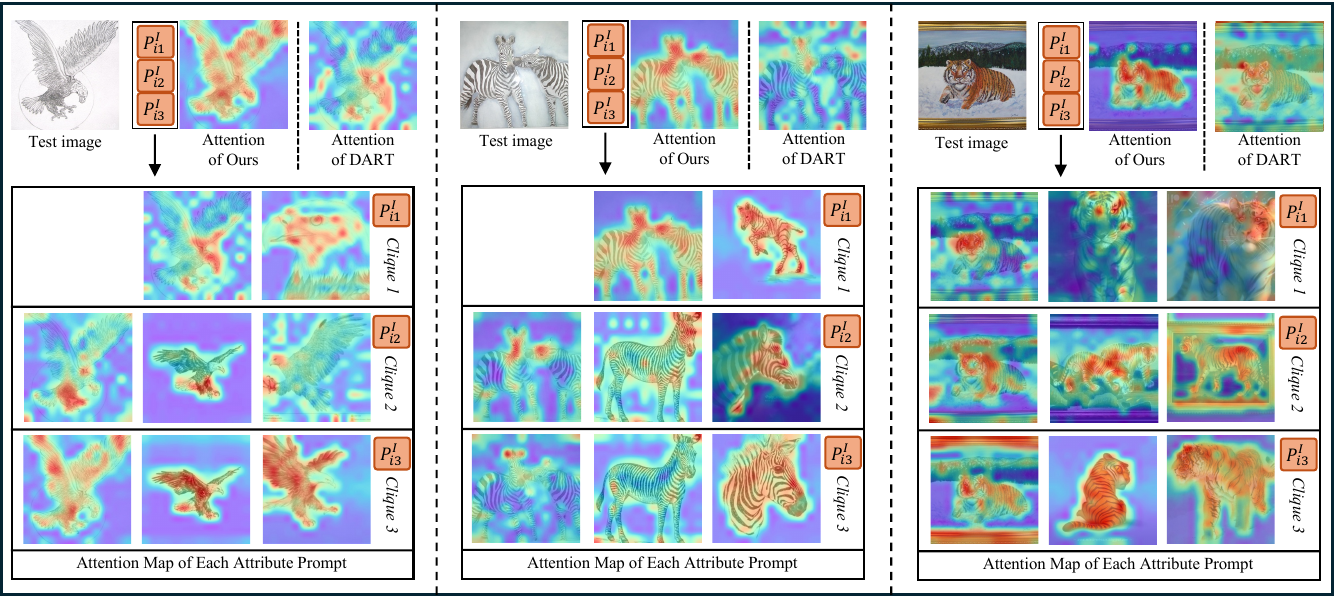}
  \vfill
  \caption{Visualization of our prompt-guided attention maps in comparison with the state-of-the-art prompt-based method.
  }
\label{fig:visualization}
\end{figure*}

\textbf{The Influence of Different Hyperparameters.} The key hyperparameters in our approach include the threshold $t$ for clique extraction, the concentration loss weight $\lambda$, the retention cache limit $L$, and the degree of the retention graph $K$. As illustrated in \cref{fig:hyper-parameters} (a), the model performance is relatively sensitive to changes in $t$. A lower $t$ results in cliques containing unrelated images, which hinders effective attribute prompt learning, whereas a higher $t$ limits the number of support samples, leading to insufficient attribute modeling.
Subsequently, we explore the impact of $\lambda$ in \cref{fig:hyper-parameters} (b). The model performance improves consistently up to $\lambda = 1.0$, as the concentration loss enhances attribute learning. However, further increasing $\lambda$ leads to overfitting, negatively affecting generalization. 
Then, we investigate the influence of $L$ in \cref{fig:hyper-parameters} (c).  We observe that as $L$ increases, the performance increases and then becomes stable, indicating that while a larger cache enhances retention, excessive stored knowledge does not yield further gains.
Finally, we study the influence of $K$ in \cref{fig:hyper-parameters} (d). Increasing $K$ up to 2 improves performance by strengthening graph connectivity and enhancing information propagation. However, beyond this point, performance declines due to the inclusion of irrelevant nodes, which introduces noise and reduces the discriminative power of the learned attributes.

\begin{table}[tb]
\centering
\small 
\setlength{\tabcolsep}{3mm} 
\renewcommand{\arraystretch}{1.2} 
\begin{tabularx}{8cm}{>{\centering\arraybackslash}X|>{\centering\arraybackslash}X|>{\centering\arraybackslash}X|>{\centering\arraybackslash}X|c}
\hline
 \multicolumn{2}{c|}{Combine $\mathcal{P}_{ik}^I$} &\multicolumn{2}{c|}{Combine $P^I$, $P^{I*}$} & ImageNet-A \\
\hline
 Mean & Concat & Mean & Concat & Acc@1\\
\hline
 \ding{51} & \ding{55} & \ding{51} & \ding{55} & 61.82\\
 \ding{51} & \ding{55} & \ding{55} & \ding{51} & 62.11\\
 \ding{55} & \ding{51} & \ding{51} & \ding{55} & 64.35\\
\hline
\rowcolor[HTML]{ECF5F9}
  \ding{55} & \ding{51} & \ding{55} & \ding{51} & \textbf{64.52}\\
\hline
\end{tabularx}
  \caption{The influence of different combination approaches.}
\label{tab:aggregation} 
\end{table}

\textbf{Influence of Different Prompt Combination Approaches.} Two primary strategies for combining multiple prompts are considered: generating a mean prompt and concatenating prompts along the token dimension, as illustrated in \cref{concat1} and in \cref{concat2}. Experimental results in Table \cref{tab:aggregation} indicate that concatenation outperforms averaging in our SCAP framework. This advantage due to concatenation preserves the full prompt information, enabling each prompt to focus on distinct image regions. In contrast, the mean prompt approach struggles to capture diverse visual patterns, leading to suboptimal performance. 

\textbf{Visualization of Supportive Cliques.} 
To demonstrate the effectiveness of our supportive clique-based attribute prompt learning approach, we visualize the attention maps in comparison with the state-of-the-art prompt-based method DART~\cite{Liu_Sun_Peng_Zhou_2024}. As illustrated in \cref{fig:visualization}, each supportive clique in our method guides the prompts to focus on specific object parts, such as the head, leg, or body. When the attribute prompts are exploited together, the obtained attention maps precisely and comprehensively capture the object information, highlighting the superior feature-learning capability of our approach over DART.
\section{Conclusion}
In this paper, we propose a novel transductive test-time adaptation approach, Supportive Clique-based Attribute Prompting (SCAP), which leverages intrinsic cross-sample shared visual attributes to enhance model adaptation during testing. Specifically, our method identifies and models the attributes within batches through supportive cliques, sets of samples exhibiting similar patterns. Furthermore, a retention module is designed to dynamically store, aggregate, and utilize knowledge from previously learned attributes, thereby improving predictions for subsequent test images. Extensive experiments on two benchmarks demonstrate the effectiveness of our method in leveraging batch-wise information. The proposed approach offers a promising direction for fully exploiting cross-sample relationships in practical transductive TTA scenarios.
 
\noindent{\textbf{Acknowledgments}}. 
 This work was supported by the grants from the National Natural Science Foundation of China (62376011, 61925201, 62132001, 62432001) and Beijing Natural Science Foundation (L247006).
{
    \small
    \bibliographystyle{ieeenat_fullname}
    \bibliography{main}

\begin{thebibliography}{57}
\providecommand{\natexlab}[1]{#1}
\providecommand{\url}[1]{\texttt{#1}}
\expandafter\ifx\csname urlstyle\endcsname\relax
  \providecommand{\doi}[1]{doi: #1}\else
  \providecommand{\doi}{doi: \begingroup \urlstyle{rm}\Url}\fi

\bibitem[Boudiaf et~al.(2022)Boudiaf, Mueller, Ben~Ayed, and Bertinetto]{boudiaf2022parameter}
Malik Boudiaf, Romain Mueller, Ismail Ben~Ayed, and Luca Bertinetto.
\newblock Parameter-free online test-time adaptation.
\newblock In \emph{CVPR}, pages 8344--8353, 2022.

\bibitem[Chen et~al.(2022)Chen, Wang, Darrell, and Ebrahimi]{chen2022contrastive}
Dian Chen, Dequan Wang, Trevor Darrell, and Sayna Ebrahimi.
\newblock Contrastive test-time adaptation.
\newblock In \emph{CVPR}, pages 295--305, 2022.

\bibitem[Cui et~al.(2024)Cui, Peng, Wang, Zhu, and Zhou]{cui2024continual}
Zhenyu Cui, Yuxin Peng, Xun Wang, Manyu Zhu, and Jiahuan Zhou.
\newblock Continual vision-language retrieval via dynamic knowledge rectification.
\newblock In \emph{AAAI}, pages 11704--11712, 2024.

\bibitem[Deng et~al.(2009)Deng, Dong, Socher, Li, Li, and Fei-Fei]{deng2009imagenet}
Jia Deng, Wei Dong, Richard Socher, Li-Jia Li, Kai Li, and Li Fei-Fei.
\newblock Imagenet: A large-scale hierarchical image database.
\newblock In \emph{CVPR}, pages 248--255. Ieee, 2009.

\bibitem[Desai and Johnson(2021)]{desai2021virtexlearningvisualrepresentations}
Karan Desai and Justin Johnson.
\newblock Virtex: Learning visual representations from textual annotations.
\newblock In \emph{CVPR}, pages 11162--11173, 2021.

\bibitem[D{\"o}bler et~al.(2023)D{\"o}bler, Marsden, and Yang]{roid}
Mario D{\"o}bler, Robert~A Marsden, and Bin Yang.
\newblock Robust mean teacher for continual and gradual test-time adaptation.
\newblock In \emph{CVPR}, pages 7704--7714, 2023.

\bibitem[Farina et~al.(2024)Farina, Franchi, Iacca, Mancini, Ricci, et~al.]{farina2024}
Matteo Farina, Gianni Franchi, Giovanni Iacca, Massimiliano Mancini, Elisa Ricci, et~al.
\newblock Frustratingly easy test-time adaptation of vision-language models.
\newblock In \emph{NeurIPS}, 2024.

\bibitem[Fei-Fei et~al.(2004)Fei-Fei, Fergus, and Perona]{fei2004learning}
Li Fei-Fei, Rob Fergus, and Pietro Perona.
\newblock Learning generative visual models from few training examples: An incremental bayesian approach tested on 101 object categories.
\newblock In \emph{CVPRW}, pages 178--178. IEEE, 2004.

\bibitem[Feng et~al.(2023)Feng, Yu, Liu, Khan, and Zuo]{feng2023diversedataaugmentationdiffusions}
Chun-Mei Feng, Kai Yu, Yong Liu, Salman Khan, and Wangmeng Zuo.
\newblock Diverse data augmentation with diffusions for effective test-time prompt tuning.
\newblock In \emph{ICCV}, pages 2704--2714, 2023.

\bibitem[Fu et~al.(2022)Fu, Du, Ding, Wang, Jiang, and Zhang]{fu2022domain}
Lihua Fu, Yubin Du, Yu Ding, Dan Wang, Hanxu Jiang, and Haitao Zhang.
\newblock Domain adaptive learning with multi-granularity features for unsupervised person re-identification.
\newblock \emph{Chinese Journal of Electronics}, 31\penalty0 (1):\penalty0 116--128, 2022.

\bibitem[Gan et~al.(2023{\natexlab{a}})Gan, Bai, Lou, Ma, Zhang, Shi, and Luo]{decoratenewcomersvisualdomain}
Yulu Gan, Yan Bai, Yihang Lou, Xianzheng Ma, Renrui Zhang, Nian Shi, and Lin Luo.
\newblock Decorate the newcomers: Visual domain prompt for continual test time adaptation.
\newblock In \emph{AAAI}, pages 7595--7603, 2023{\natexlab{a}}.

\bibitem[Gan et~al.(2023{\natexlab{b}})Gan, Bai, Lou, Ma, Zhang, Shi, and Luo]{gan2023decorate}
Yulu Gan, Yan Bai, Yihang Lou, Xianzheng Ma, Renrui Zhang, Nian Shi, and Lin Luo.
\newblock Decorate the newcomers: Visual domain prompt for continual test time adaptation.
\newblock In \emph{AAAI}, pages 7595--7603, 2023{\natexlab{b}}.

\bibitem[Gong et~al.(2024)Gong, Kim, Lee, Chottananurak, and Lee]{sotta2023}
Taesik Gong, Yewon Kim, Taeckyung Lee, Sorn Chottananurak, and Sung-Ju Lee.
\newblock Sotta: Robust test-time adaptation on noisy data streams.
\newblock \emph{NeurIPS}, 36, 2024.

\bibitem[Hakim et~al.(2024)Hakim, Osowiechi, Noori, Cheraghalikhani, Bahri, Yazdanpanah, Ayed, and Desrosiers]{hakim2024clipartt}
Gustavo Adolfo~Vargas Hakim, David Osowiechi, Mehrdad Noori, Milad Cheraghalikhani, Ali Bahri, Moslem Yazdanpanah, Ismail~Ben Ayed, and Christian Desrosiers.
\newblock Clipartt: Light-weight adaptation of clip to new domains at test time.
\newblock \emph{arXiv preprint arXiv:2405.00754}, 2024.

\bibitem[Hendrycks et~al.(2021{\natexlab{a}})Hendrycks, Basart, Mu, Kadavath, Wang, Dorundo, Desai, Zhu, Parajuli, Guo, et~al.]{hendrycks2021facesrobustnesscriticalanalysis}
Dan Hendrycks, Steven Basart, Norman Mu, Saurav Kadavath, Frank Wang, Evan Dorundo, Rahul Desai, Tyler Zhu, Samyak Parajuli, Mike Guo, et~al.
\newblock The many faces of robustness: A critical analysis of out-of-distribution generalization.
\newblock In \emph{ICCV}, pages 8340--8349, 2021{\natexlab{a}}.

\bibitem[Hendrycks et~al.(2021{\natexlab{b}})Hendrycks, Zhao, Basart, Steinhardt, and Song]{hendrycks2021natural}
Dan Hendrycks, Kevin Zhao, Steven Basart, Jacob Steinhardt, and Dawn Song.
\newblock Natural adversarial examples.
\newblock In \emph{CVPR}, pages 15262--15271, 2021{\natexlab{b}}.

\bibitem[Hu et~al.(2021)Hu, Uzunbas, Chen, Wang, Shah, Nevatia, and Lim]{hu2021mixnorm}
Xuefeng Hu, Gokhan Uzunbas, Sirius Chen, Rui Wang, Ashish Shah, Ram Nevatia, and Ser-Nam Lim.
\newblock Mixnorm: Test-time adaptation through online normalization estimation.
\newblock \emph{arXiv preprint arXiv:2110.11478}, 2021.

\bibitem[Hu et~al.(2022)Hu, Gan, Wang, Yang, Liu, Lu, and Wang]{hu2022scalingvisionlanguagepretrainingimage}
Xiaowei Hu, Zhe Gan, Jianfeng Wang, Zhengyuan Yang, Zicheng Liu, Yumao Lu, and Lijuan Wang.
\newblock Scaling up vision-language pre-training for image captioning.
\newblock In \emph{CVPR}, pages 17980--17989, 2022.

\bibitem[Jia et~al.(2021)Jia, Yang, Xia, Chen, Parekh, Pham, Le, Sung, Li, and Duerig]{scalingvisualvisionlanguagerepresentation}
Chao Jia, Yinfei Yang, Ye Xia, Yi-Ting Chen, Zarana Parekh, Hieu Pham, Quoc Le, Yun-Hsuan Sung, Zhen Li, and Tom Duerig.
\newblock Scaling up visual and vision-language representation learning with noisy text supervision.
\newblock In \emph{ICML}, pages 4904--4916. PMLR, 2021.

\bibitem[Karmanov et~al.(2024)Karmanov, Guan, Lu, El~Saddik, and Xing]{karmanov2024}
Adilbek Karmanov, Dayan Guan, Shijian Lu, Abdulmotaleb El~Saddik, and Eric Xing.
\newblock Efficient test-time adaptation of vision-language models.
\newblock In \emph{CVPR}, pages 14162--14171, 2024.

\bibitem[Khattak et~al.(2023)Khattak, Rasheed, Maaz, Khan, and Khan]{maple}
Muhammad~Uzair Khattak, Hanoona Rasheed, Muhammad Maaz, Salman Khan, and Fahad~Shahbaz Khan.
\newblock Maple: Multi-modal prompt learning.
\newblock In \emph{CVPR}, pages 19113--19122, 2023.

\bibitem[Li et~al.(2024)Li, Xu, Peng, and Zhou]{li2024exemplar}
Qiwei Li, Kunlun Xu, Yuxin Peng, and Jiahuan Zhou.
\newblock Exemplar-free lifelong person re-identification via prompt-guided adaptive knowledge consolidation.
\newblock \emph{IJCV}, 132\penalty0 (11):\penalty0 4850--4865, 2024.

\bibitem[Li et~al.(2023)Li, Lian, Lu, Bai, Chen, and Wang]{li2023graphadaptertuningvisionlanguagemodels}
Xin Li, Dongze Lian, Zhihe Lu, Jiawang Bai, Zhibo Chen, and Xinchao Wang.
\newblock Graphadapter: Tuning vision-language models with dual knowledge graph.
\newblock \emph{NeurIPS}, 36:\penalty0 13448--13466, 2023.

\bibitem[Liu et~al.(2024{\natexlab{a}})Liu, Peng, and Zhou]{liu2024insvp}
Zichen Liu, Yuxin Peng, and Jiahuan Zhou.
\newblock Insvp: Efficient instance visual prompting from image itself.
\newblock In \emph{Proceedings of the 32nd ACM International Conference on Multimedia}, pages 6443--6452, 2024{\natexlab{a}}.

\bibitem[Liu et~al.(2024{\natexlab{b}})Liu, Sun, Peng, and Zhou]{Liu_Sun_Peng_Zhou_2024}
Zichen Liu, Hongbo Sun, Yuxin Peng, and Jiahuan Zhou.
\newblock Dart: dual-modal adaptive online prompting and knowledge retention for test-time adaptation.
\newblock In \emph{AAAI}, pages 14106--14114, 2024{\natexlab{b}}.

\bibitem[Ma et~al.(2024)Ma, Zhang, Guo, and Xu]{ma2024swapprompt}
Xiaosong Ma, Jie Zhang, Song Guo, and Wenchao Xu.
\newblock Swapprompt: Test-time prompt adaptation for vision-language models.
\newblock \emph{NeurIPS}, 36, 2024.

\bibitem[Maji et~al.(2013)Maji, Rahtu, Kannala, Blaschko, and Vedaldi]{maji2013finegrainedvisualclassificationaircraft}
Subhransu Maji, Esa Rahtu, Juho Kannala, Matthew Blaschko, and Andrea Vedaldi.
\newblock Fine-grained visual classification of aircraft.
\newblock \emph{arXiv preprint arXiv:1306.5151}, 2013.

\bibitem[Marsden et~al.(2024)Marsden, D{\"o}bler, and Yang]{marsden2023universaltesttimeadaptationweight}
Robert~A Marsden, Mario D{\"o}bler, and Bin Yang.
\newblock Universal test-time adaptation through weight ensembling, diversity weighting, and prior correction.
\newblock In \emph{WACV}, pages 2555--2565, 2024.

\bibitem[Nilsback and Zisserman(2008)]{nilsback2008automated}
Maria-Elena Nilsback and Andrew Zisserman.
\newblock Automated flower classification over a large number of classes.
\newblock In \emph{2008 Sixth Indian conference on computer vision, graphics \& image processing}, pages 722--729. IEEE, 2008.

\bibitem[Parkhi et~al.(2012)Parkhi, Vedaldi, Zisserman, and Jawahar]{parkhi2012cats}
Omkar~M Parkhi, Andrea Vedaldi, Andrew Zisserman, and CV Jawahar.
\newblock Cats and dogs.
\newblock In \emph{CVPR}, pages 3498--3505. IEEE, 2012.

\bibitem[Press et~al.(2023)Press, Schneider, K{\"u}mmerer, and Bethge]{press2024rdumb}
Ori Press, Steffen Schneider, Matthias K{\"u}mmerer, and Matthias Bethge.
\newblock Rdumb: A simple approach that questions our progress in continual test-time adaptation.
\newblock \emph{NeurIPS}, 36:\penalty0 39915--39935, 2023.

\bibitem[Radford et~al.(2021{\natexlab{a}})Radford, Kim, Hallacy, Ramesh, Goh, Agarwal, Sastry, Askell, Mishkin, Clark, et~al.]{radford2021}
Alec Radford, Jong~Wook Kim, Chris Hallacy, Aditya Ramesh, Gabriel Goh, Sandhini Agarwal, Girish Sastry, Amanda Askell, Pamela Mishkin, Jack Clark, et~al.
\newblock Learning transferable visual models from natural language supervision.
\newblock In \emph{ICML}, pages 8748--8763. PMLR, 2021{\natexlab{a}}.

\bibitem[Radford et~al.(2021{\natexlab{b}})Radford, Kim, Hallacy, Ramesh, Goh, Agarwal, Sastry, Askell, Mishkin, Clark, et~al.]{radford2021learningtransferablevisualmodels}
Alec Radford, Jong~Wook Kim, Chris Hallacy, Aditya Ramesh, Gabriel Goh, Sandhini Agarwal, Girish Sastry, Amanda Askell, Pamela Mishkin, Jack Clark, et~al.
\newblock Learning transferable visual models from natural language supervision.
\newblock In \emph{ICML}, pages 8748--8763. PmLR, 2021{\natexlab{b}}.

\bibitem[Recht et~al.(2019)Recht, Roelofs, Schmidt, and Shankar]{recht2019imagenetclassifiersgeneralizeimagenet}
Benjamin Recht, Rebecca Roelofs, Ludwig Schmidt, and Vaishaal Shankar.
\newblock Do imagenet classifiers generalize to imagenet?
\newblock In \emph{ICML}, pages 5389--5400. PMLR, 2019.

\bibitem[Schneider et~al.(2020)Schneider, Rusak, Eck, Bringmann, Brendel, and Bethge]{schneider2020improving}
Steffen Schneider, Evgenia Rusak, Luisa Eck, Oliver Bringmann, Wieland Brendel, and Matthias Bethge.
\newblock Improving robustness against common corruptions by covariate shift adaptation.
\newblock \emph{NeurIPS}, 33:\penalty0 11539--11551, 2020.

\bibitem[Shu et~al.(2022)Shu, Nie, Huang, Yu, Goldstein, Anandkumar, and Xiao]{shu2022}
Manli Shu, Weili Nie, De-An Huang, Zhiding Yu, Tom Goldstein, Anima Anandkumar, and Chaowei Xiao.
\newblock Test-time prompt tuning for zero-shot generalization in vision-language models.
\newblock \emph{NeurIPS}, 35:\penalty0 14274--14289, 2022.

\bibitem[Song et~al.(2023)Song, Lee, Kweon, and Choi]{song2023ecottamemoryefficientcontinualtesttime}
Junha Song, Jungsoo Lee, In~So Kweon, and Sungha Choi.
\newblock Ecotta: Memory-efficient continual test-time adaptation via self-distilled regularization.
\newblock In \emph{CVPR}, pages 11920--11929, 2023.

\bibitem[Wang et~al.()Wang, Shelhamer, Liu, Olshausen, and Darrell]{wang2021tentfullytesttimeadaptation}
Dequan Wang, Evan Shelhamer, Shaoteng Liu, Bruno Olshausen, and Trevor Darrell.
\newblock Tent: Fully test-time adaptation by entropy minimization.
\newblock In \emph{ICLR}.

\bibitem[Wang et~al.(2020)Wang, Shelhamer, Liu, Olshausen, and Darrell]{wang2020tent}
Dequan Wang, Evan Shelhamer, Shaoteng Liu, Bruno Olshausen, and Trevor Darrell.
\newblock Tent: Fully test-time adaptation by entropy minimization.
\newblock \emph{arXiv preprint arXiv:2006.10726}, 2020.

\bibitem[Wang et~al.(2019)Wang, Ge, Lipton, and Xing]{wang2019learningrobustglobalrepresentations}
Haohan Wang, Songwei Ge, Zachary Lipton, and Eric~P Xing.
\newblock Learning robust global representations by penalizing local predictive power.
\newblock \emph{NeurIPS}, 32, 2019.

\bibitem[Wang et~al.(2022)Wang, Fink, Van~Gool, and Dai]{wang2022continual}
Qin Wang, Olga Fink, Luc Van~Gool, and Dengxin Dai.
\newblock Continual test-time domain adaptation.
\newblock In \emph{CVPR}, pages 7201--7211, 2022.

\bibitem[Xu et~al.(2024{\natexlab{a}})Xu, Zhang, Li, Peng, and Zhou]{xu2024mitigate}
Kunlun Xu, Haozhuo Zhang, Yu Li, Yuxin Peng, and Jiahuan Zhou.
\newblock Mitigate catastrophic remembering via continual knowledge purification for noisy lifelong person re-identification.
\newblock In \emph{ACM MM}, pages 5790--5799, 2024{\natexlab{a}}.

\bibitem[Xu et~al.(2024{\natexlab{b}})Xu, Zou, Peng, and Zhou]{xu2024distribution}
Kunlun Xu, Xu Zou, Yuxin Peng, and Jiahuan Zhou.
\newblock Distribution-aware knowledge prototyping for non-exemplar lifelong person re-identification.
\newblock In \emph{CVPR}, pages 16604--16613, 2024{\natexlab{b}}.

\bibitem[Xu et~al.(2024{\natexlab{c}})Xu, Zou, and Zhou]{xu2024lstkc}
Kunlun Xu, Xu Zou, and Jiahuan Zhou.
\newblock Lstkc: Long short-term knowledge consolidation for lifelong person re-identification.
\newblock In \emph{AAAI}, pages 16202--16210, 2024{\natexlab{c}}.

\bibitem[Yang et~al.(2022)Yang, Duan, Tran, Xu, Chanda, Chen, Zeng, Chilimbi, and Huang]{yang2022visionlanguagepretrainingtriplecontrastive}
Jinyu Yang, Jiali Duan, Son Tran, Yi Xu, Sampath Chanda, Liqun Chen, Belinda Zeng, Trishul Chilimbi, and Junzhou Huang.
\newblock Vision-language pre-training with triple contrastive learning.
\newblock In \emph{CVPR}, pages 15671--15680, 2022.

\bibitem[Ye et~al.(2022)Ye, He, and Peng]{ye2022unsupervised}
Zhaoda Ye, Xiangteng He, and Yuxin Peng.
\newblock Unsupervised cross-media hashing learning via knowledge graph.
\newblock \emph{Chinese Journal of Electronics}, 31\penalty0 (6):\penalty0 1081--1091, 2022.

\bibitem[Yoon et~al.(2024)Yoon, Yoon, Tee, Hasegawa-Johnson, Li, and Yoo]{yoon2024c}
Hee~Suk Yoon, Eunseop Yoon, Joshua Tian~Jin Tee, Mark Hasegawa-Johnson, Yingzhen Li, and Chang~D Yoo.
\newblock C-tpt: Calibrated test-time prompt tuning for vision-language models via text feature dispersion.
\newblock \emph{arXiv preprint arXiv:2403.14119}, 2024.

\bibitem[Yu et~al.(2023)Yu, Lu, Jin, Chen, and Wang]{yu2023taskresidualtuningvisionlanguage}
Tao Yu, Zhihe Lu, Xin Jin, Zhibo Chen, and Xinchao Wang.
\newblock Task residual for tuning vision-language models.
\newblock In \emph{CVPR}, pages 10899--10909, 2023.

\bibitem[Zanella and Ben~Ayed(2024)]{Zanella_2024_CVPR}
Maxime Zanella and Ismail Ben~Ayed.
\newblock On the test-time zero-shot generalization of vision-language models: Do we really need prompt learning?
\newblock In \emph{CVPR}, pages 23783--23793, 2024.

\bibitem[Zhang et~al.(2024{\natexlab{a}})Zhang, Zhou, and Li]{Adaprompt}
Ding-Chu Zhang, Zhi Zhou, and Yu-Feng Li.
\newblock Robust test-time adaptation for zero-shot prompt tuning.
\newblock In \emph{AAAI}, pages 16714--16722, 2024{\natexlab{a}}.

\bibitem[Zhang et~al.(2022)Zhang, Zhang, Fang, Gao, Li, Dai, Qiao, and Li]{zhang2022tip}
Renrui Zhang, Wei Zhang, Rongyao Fang, Peng Gao, Kunchang Li, Jifeng Dai, Yu Qiao, and Hongsheng Li.
\newblock Tip-adapter: Training-free adaption of clip for few-shot classification.
\newblock In \emph{ECCV}, pages 493--510. Springer, 2022.

\bibitem[Zhang et~al.(2024{\natexlab{b}})Zhang, Haihong, and Song]{zhang2024fscil}
Ruru Zhang, E Haihong, and Meina Song.
\newblock Fscil-eaca: Few-shot class-incremental learning network based on embedding augmentation and classifier adaptation for image classification.
\newblock \emph{Chinese Journal of Electronics}, 33\penalty0 (1):\penalty0 139--152, 2024{\natexlab{b}}.

\bibitem[Zhou et~al.(2003)Zhou, Bousquet, Lal, Weston, and Sch{\"o}lkopf]{graph}
Dengyong Zhou, Olivier Bousquet, Thomas Lal, Jason Weston, and Bernhard Sch{\"o}lkopf.
\newblock Learning with local and global consistency.
\newblock \emph{NeurIPS}, 16, 2003.

\bibitem[Zhou et~al.(2017)Zhou, Yu, Tang, and Wu]{zhou2017efficient}
Jiahuan Zhou, Pei Yu, Wei Tang, and Ying Wu.
\newblock Efficient online local metric adaptation via negative samples for person re-identification.
\newblock In \emph{ICCV}, pages 2420--2428, 2017.

\bibitem[Zhou et~al.(2020)Zhou, Su, and Wu]{zhou2020online}
Jiahuan Zhou, Bing Su, and Ying Wu.
\newblock Online joint multi-metric adaptation from frequent sharing-subset mining for person re-identification.
\newblock In \emph{CVPR}, pages 2909--2918, 2020.

\bibitem[Zhou et~al.(2022{\natexlab{a}})Zhou, Yang, Loy, and Liu]{zhou2022conditional}
Kaiyang Zhou, Jingkang Yang, Chen~Change Loy, and Ziwei Liu.
\newblock Conditional prompt learning for vision-language models.
\newblock In \emph{CVPR}, pages 16816--16825, 2022{\natexlab{a}}.

\bibitem[Zhou et~al.(2022{\natexlab{b}})Zhou, Yang, Loy, and Liu]{zhou2022learning}
Kaiyang Zhou, Jingkang Yang, Chen~Change Loy, and Ziwei Liu.
\newblock Learning to prompt for vision-language models.
\newblock \emph{IJCV}, 130\penalty0 (9):\penalty0 2337--2348, 2022{\natexlab{b}}.

\end{thebibliography}
}


\end{document}


\clearpage
\setcounter{page}{1}
\maketitlesupplementary

\section{Benchmark Details}

Here, we provide the sub-dataset details of the out-of-distribution (OOD) and cross-domain benchmarks exploited in this paper.

\subsection{Details on Datasets in the OOD Benchmark}

Our OOD benchmark consists of four out-of-distribution datasets derived from ImageNet \cite{deng2009imagenet}.

\begin{itemize}[left=0pt]
    \item \textbf{ImageNet-A} \cite{hendrycks2021natural} contains 7,500 images of 200 classes that are naturally perturbed and misclassified by ResNet-50 \cite{he2016deep} in ImageNet.
    \item \textbf{ImageNet-R} \cite{hendrycks2021facesrobustnesscriticalanalysis}includes 30,000 images covering 200 classes across 16 artistic and stylistic domains, such as cartoons, graffiti, and sketches, posing significant challenges due to their diverse visual transformations.
    \item \textbf{ImageNet-Sketch} \cite{wang2019learningrobustglobalrepresentations} contains 50,000 images of 1,000 categories. The distribution of ImageNet-Sketch differs greatly from the pre-training data of CLIP since it only contains black-and-white sketches. Thus, it has been a challenging dataset for TTA.
    \item \textbf{ImageNet-V2} \cite{recht2019imagenetclassifiersgeneralizeimagenet} comprises 10,000 images across 1,000 classes, sampled a decade after the original ImageNet dataset. It presents a naturally evolved distribution shift, making it a realistic benchmark for evaluating generalization performance..
\end{itemize}

\begin{table}[h!]
\centering
\begin{tabular}{lcc}
\toprule
\textbf{OOD Datasets} & \textbf{Size} & \textbf{Number of Classes} \\
\midrule
ImageNet-A           & 7,500 & 200   \\
ImageNet-R           & 30,000& 200    \\
ImageNet-Sketch      & 50,000& 1,000   \\
ImageNet-V2          & 10,000& 1,000    \\

\bottomrule
\end{tabular}
\caption{Overview of datasets in OOD benchmark}
\label{tab:datasets}
\end{table}

\begin{table}[h!]
\centering
\begin{tabular}{lcc}
\toprule
\textbf{Cross-Domain} & \textbf{Size} & \textbf{Number of Classes} \\
\midrule
Aircraft           & 3,333 & 100   \\
Caltech101           & 2,465 & 100    \\
Flower102      & 2,463 & 102   \\
Pets          & 3,669 & 37    \\

\bottomrule
\end{tabular}
\caption{Overview of datasets in Cross-domain benchmark.}
\label{tab:datasets}
\end{table}

\begin{figure}[!t]
\centering
\includegraphics[width=\linewidth]{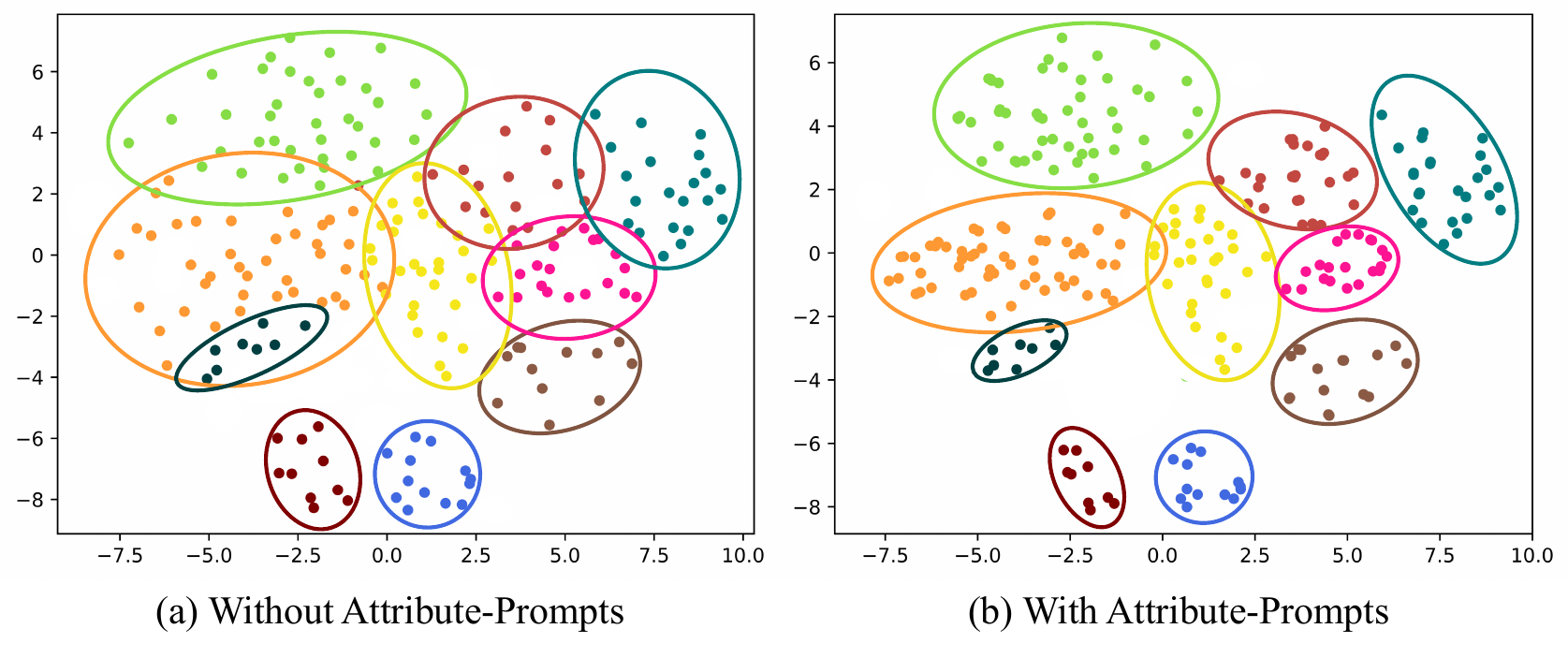}
  \vfill
  \caption{Visualization of the distribution of image features with and without the learned Attribute Prompts.}
\label{fig:prompt_visual}
\end{figure}

\subsection{Details on Datasets in Cross-domain Benchmark}

The cross-domain benchmark comprises four datasets from distinct visual domains, providing a diverse evaluation setting for assessing model generalization.
\begin{itemize}[left=0pt]
    \item \textbf{Aircraft} \cite{maji2013finegrainedvisualclassificationaircraft}  consists of 10,200 images representing 102 different aircraft model variants, each with 100 images. The dataset predominantly features airplanes and poses challenges in fine-grained visual classification.
    \item \textbf{Caltech101} \cite{fei2004learning}  includes 2,465 images spanning 101 object categories along with a background class. Each category contains 40 to 800 images, with most classes averaging around 50 samples, making it a benchmark for object recognition tasks.
    \item \textbf{Flower102} \cite{nilsback2008automated} comprises 2,463 images of 102 flower species. Due to the significant domain gap between flowers and standard pretraining datasets, this dataset serves as a robust test for domain generalization.
    \item \textbf{Pets} \cite{parkhi2012cats} contains images of cats and dogs across 37 different breeds. The dataset exhibits high intra-class variations in scale, pose, and lighting, making it challenging for fine-grained classification tasks.
\end{itemize}

\begin{figure*}[hbtp]
\centering
\includegraphics[width=\textwidth]{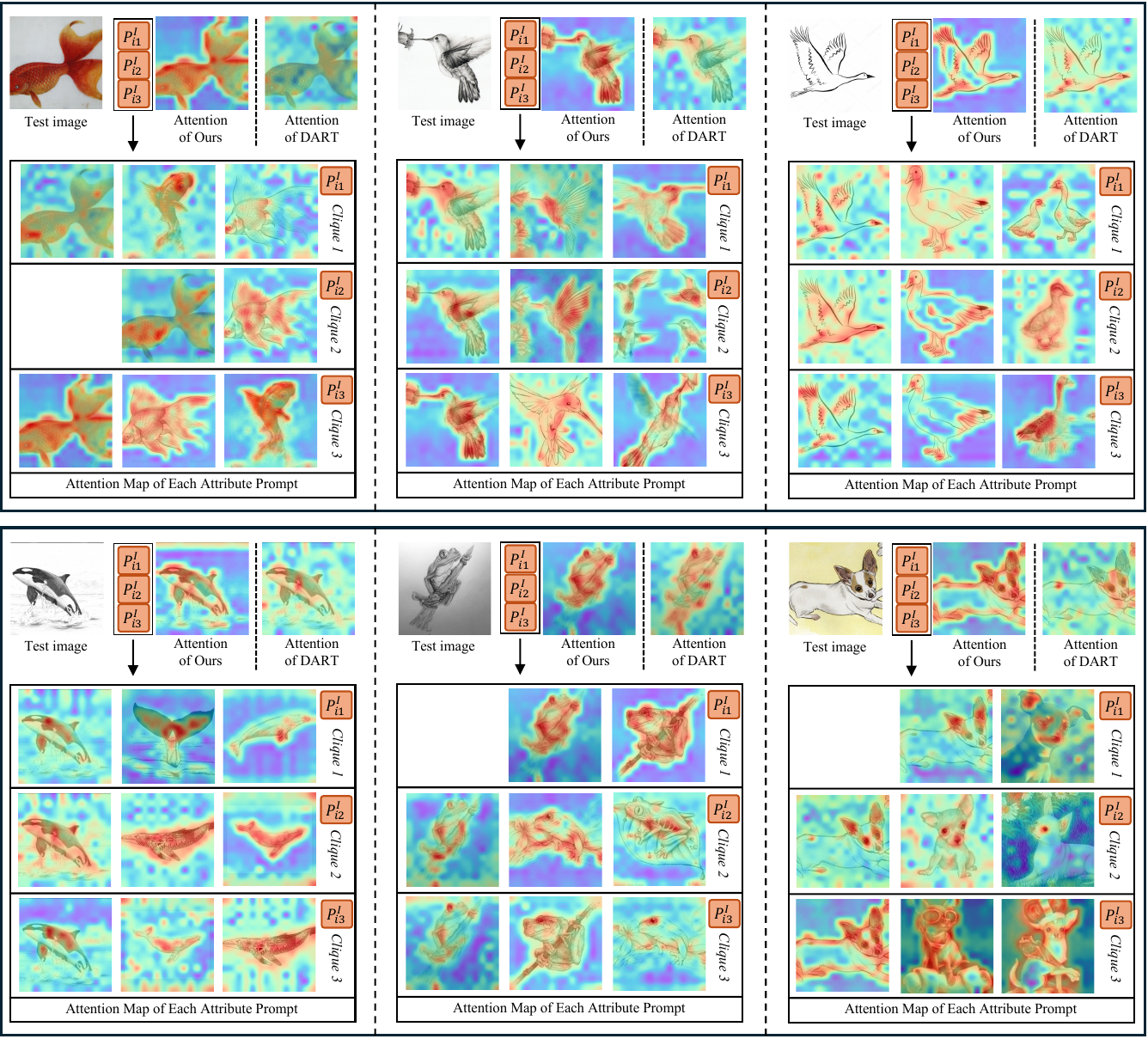}
  \vfill
  \caption{Additional visualization results of the attention maps.
  }
\label{fig:visual2}
\end{figure*}
\renewcommand\arraystretch{0.95}
\begin{table*}[ht]
  \centering
  \setlength{\tabcolsep}{0.8mm}{
    \begin{tabular}{l*{13}c}
      \toprule
      Method & Publication & Cars & SUN397 & Aircraft & EuroSAT & Food101 & Pets & Flower102 & Caltech101 & DTD & UCF101 & Average \\
      \midrule
      TDA        &  CVPR 2024  & 67.28    & 67.62	&23.91 & 58.00	&86.14	&88.63	&71.42	&94.24 &47.40	&70.66	&67.53\\
      \midrule
      \textbf{SCAP}   &  \textbf{This Paper}& \textbf{69.25} & 66.41	& \textbf{25.44}	& \textbf{58.62} & \textbf{86.58}	& \textbf{90.27}	& \textbf{71.65}	& \textbf{94.42}	&\textbf{47.79}	&68.36	& \textbf{67.88} \\
      \bottomrule
    \end{tabular}
  } 
  \caption{\textbf{Additional Results on the domain-shift datasets.} SCAP compared with the second-highest TDA.}
  \label{tab:crossdomain}
\end{table*}

\section{The Effectiveness of Attribute Prompts} 

To demonstrate the effectiveness of our supportive clique-based attribute prompting, we visualize the embedding distributions of images from different classes in Figure \ref{fig:prompt_visual}, where with and without the corresponding Attribute Prompts setting are conducted. As illustrated in Figure \ref{fig:prompt_visual}, incorporating Attribute Prompts improves intra-class compactness and inter-class discrimination. Notably, several previously ambiguous samples near class boundaries shift closer to their respective class centers, thereby enhancing the overall quality of the learned image representations.
This improvement is attributed to our proposed \textit{Concentration Loss}, which, together with \textit{Entropy Loss}, encourages samples within the same supportive clique to leverage shared attributes, leading to more tightly clustered and semantically consistent features.

\section{Visualization of Attention Maps} 

In \cref{fig:visual2}, we provide additional visualizations of the attention maps in comparison with DART~\cite{Liu_Sun_Peng_Zhou_2024}. The results demonstrate that each of our \textit{visual attribute prompts} guides CLIP to attend to the specific attributes shared among images within a supportive clique, thereby enabling fine-grained attribute-based prompt learning. By aggregating all relevant \textit{visual attribute prompts} associated with a given test image, our approach effectively directs attention toward the most salient attributes, leading to enhanced feature extraction and improved classification performance. In contrast, the attention maps generated by DART often exhibit dispersed or incomplete attention, failing to sufficiently capture critical object attributes. These findings highlight the superior prompt-learning capability of SCAP in accurately and comprehensively leveraging visual information.

\section{Additional Results on the domain-shift datasets} 
In \cref{tab:crossdomain}, we report results on six additional datasets alongside the four datasets in our main paper from the cross-domain benchmark. SCAP consistently outperforms the second-best method, TDA, on eight out of ten datasets, achieving an average accuracy improvement of \textbf{0.35\%}. These results further demonstrate SCAP's effectiveness in adapting to diverse domain shifts, highlighting its robustness in transductive TTA scenarios.

{
    \small
    \bibliographystyle{ieeenat_fullname}
    \bibliography{main}
}